\documentclass[a4paper, 10pt, conference]{ieeeconf}      
                                                          
\usepackage{FG2020}

\usepackage{graphicx}
\usepackage{color}
\usepackage{xcolor}
\usepackage{amssymb}
\usepackage{comment}

\usepackage{amsmath,amssymb,amsfonts}
\usepackage{algorithmic}
\usepackage{graphics}
\usepackage{textcomp}
\usepackage[export]{adjustbox}
\usepackage{multicol}
\usepackage{tabularx}
\usepackage{color}

\usepackage[bookmarks=false]{hyperref}
\hypersetup{
    colorlinks=true,
    linkcolor=blue,
    filecolor=magenta,      
    urlcolor=red,
}
\FGfinalcopy

\IEEEoverridecommandlockouts                             

\def\FGPaperID{****} 

\title{\LARGE \bf
Analysing Affective Behavior \\ in the First ABAW 2020 Competition
}

\author{\parbox{16cm}{\centering
    {\large Dimitrios Kollias $^1$, Attila Schulc $^2$, Elnar Hajiyev $^2$ and Stefanos Zafeiriou $^1$}\\
    {\normalsize
    $^1$ Department of Computing, Imperial College London, UK\\
    $^2$ Realeyes}}
}

\begin{document}

\ifFGfinal
\thispagestyle{empty}
\pagestyle{empty}
\else
\author{Anonymous FG2020 submission\\ Paper ID \FGPaperID \\}
\pagestyle{plain}
\fi
\maketitle

\begin{abstract}
The Affective Behavior Analysis in-the-wild (ABAW) 2020 Competition is the first Competition aiming at automatic analysis of the three main behavior tasks of valence-arousal estimation, basic expression recognition and action unit detection. 
It is split into three Challenges, each one addressing a respective behavior task. 
For the Challenges, we provide a common benchmark database, Aff-Wild2, which is a large scale in-the-wild database and the first one annotated for all these three tasks.
In this paper, we describe this Competition, to be held in conjunction with the IEEE Conference on Face and Gesture Recognition, May 2020, in Buenos Aires, Argentina. We present the three Challenges, with the utilized Competition corpora. We outline the evaluation metrics, present both the baseline system and the top-3 performing teams' methodologies per Challenge and finally present their obtained results. More information regarding the Competition, the leaderboard of each Challenge and details for accessing the utilized database, are provided in the Competition site: \url{http://ibug.doc.ic.ac.uk/resources/fg-2020-competition-affective-behavior-analysis}.

\end{abstract}




\section{Introduction}

The proposed Competition tackles the problem of affective behavior analysis in-the-wild, which is a major targeted characteristic of human computer interaction systems used in real life applications. The current $5^{th}$ societal revolution aims at merging the physical and cyber spaces, providing services that contribute to people's well-being. The target is to create machines and robots that are capable of understanding people's feelings, emotions and behaviors; thus, being able to interact in a 'human-centered' and engaging  manner with them, and effectively serving them as their digital assistants.    

Affective behavior analysis in diverse environments, i.e., in-the-wild, will have a positive societal impact, by helping machines and robots to interact with and assist people in a natural way. Through human affect recognition, the reactions of the machine, or robot, will be consistent with people's emotions \cite{kollias2015interweaving}; their verbal and non-verbal interactions will be positively received by humans. This will not be dependent on human's  age, sex, ethnicity, educational, level, profession, or social position. A great improvement in generating trust, understanding and closeness between humans and machines in everyday societal environments will be achieved through development of intelligent systems able to analyze human behaviors in-the-wild.

\begin{figure}[h]
\centering
\adjincludegraphics[height=5.3cm]{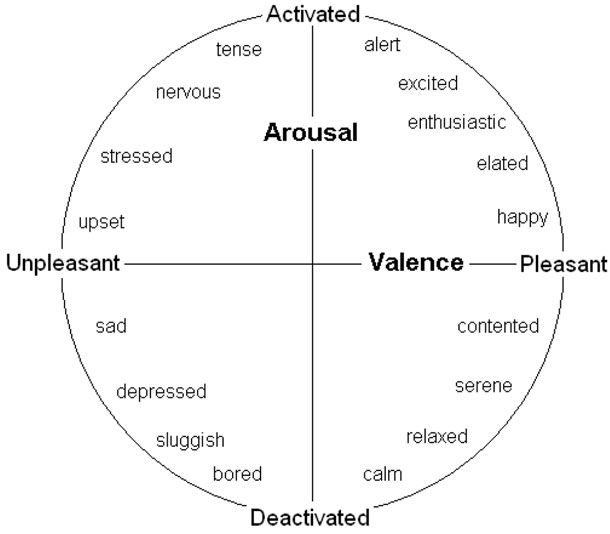}
\caption{The 2D Valence-Arousal Space}
\label{va-space}
\end{figure}

Representing human emotions has been a basic topic of research in psychology. The most frequently used emotion representation is the categorical one, including the seven basic categories, i.e., Anger, Disgust, Fear, Happiness, Sadness, Surprise and Neutral \cite{ekman2003darwin}. Many research problems \cite{zhou2019exploring,ali2019facial} have focused on this representation. Discrete emotion representation can also be described in terms of the Facial Action Coding System (FACS) model, in which all possible facial actions are described in terms of Action Units (AUs) \cite{ekman2002facial}. Many automatic methodologies \cite{he2017multi,batista2017aumpnet} have been proposed to recognise this representation. 
Finally, the dimensional model of affect \cite{whissel1989dictionary,russell1978evidence} has been proposed as a means to distinguish between subtly different displays of affect and encode small changes in the intensity of each emotion on a continuous scale. The 2-D Valence and Arousal (VA) Space (valence shows how positive or negative an emotional state is, whereas arousal shows how passive or active it is) is the most usual dimensional emotion representation, depicted in Figure \ref{va-space}. Many approaches \cite{chen2019efficient,kollias2019exploiting,kollias2018old} have been developed to automatically recognise this representation. 

The Competition will contribute to advancing the related technical state-of-the-art, by targeting, for the first time, dimensional (in terms of valence and arousal), categorical (in terms of the seven basic emotions) and facial action unit analysis and recognition. The Competition is split into three Challenges, namely Valence-Arousal estimation Challenge, Seven Basic Expression Classification Challenge and Eight Action Unit Detection Challenge. All these Challenges are based, for the first time, on the same large in-the-wild audiovisual database, the Aff-Wild2 \cite{kolliasexpression,kollias2018aff2,kollias2018multi}, which contains annotations for all these tasks. 

The remainder of this paper is organised as follows. We introduce the Competition corpora in Section \ref{corpora}, the Competition evaluation metrics in Section \ref{metrics}, the developed baseline and top-3 (in terms of performance) team's systems per Challenge, along with the obtained results in Section \ref{baseline}, before concluding in Section \ref{conclusion}.


\section{Competition Corpora}\label{corpora}


The First Affective Behavior Analysis in-the-wild (ABAW) Competition relies on the Aff-Wild2 database \cite{kolliasexpression,kollias2018aff2,kollias2018multi}. Aff-Wild2 is the first ever database annotated for all three main behavior tasks: valence-arousal estimation, action unit detection and basic expression classification. These three tasks form the three Challenges of this Competition. In the following, we provide a short overview of each Challenge's dataset and refer the reader to the original work for a more complete description. Finally, we describe the pre-processing steps that we carried out for cropping and aligning the images of Aff-Wild2. The cropped and aligned images have been provided to all Competition participating teams. Additionally, they were  utilized in our baseline experiments. 

\subsection{Aff-Wild2: Valence-Arousal Annotation}

Aff-Wild2 consists of $545$ videos with $2,786,201$ frames. Sixteen of these videos display two subjects (both have been annotated). Aff-Wild2 is currently the largest (and audiovisual) in-the-wild database annotated for valence and arousal. All videos have been collected from YouTube. Aff-Wild2 is an extension of Aff-Wild \cite{kollias2018deep,zafeiriou,zafeiriou1}; 260 more YouTube videos, with $1,413,000$ frames, have been added to Aff-Wild. 
Aff-Wild was the first large scale, captured in-the-wild, dimensionally annotated database, containing 298 YouTube videos that display subjects reacting to a variety of stimuli. 


Aff-Wild2 shows both subtle and extreme human behaviours in real-world settings. The total number of subjects in Aff-Wild2 is 458; 279 of them are males and 179 females.

\begin{figure}[h]
\centering
\adjincludegraphics[height=5.6cm]{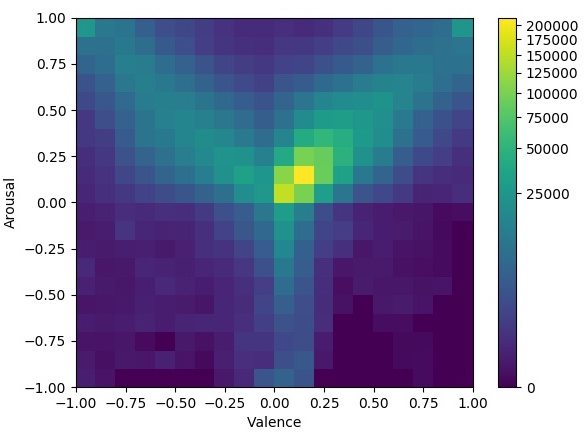}
\caption{2D Valence-Arousal Histogram of Aff-Wild2}
\label{va_annot}
\end{figure}

Four experts annotated Aff-Wild2 with respect to valence and arousal, using the method proposed in \cite{cowie2000feeltrace}. The annotators watched each video and provided their (frame-by-frame) annotations through a joystick. A time-continuous annotation was generated for each affect dimension. Valence and arousal values range continuously in $[-1,1]$.  The final label  values  were  the  mean  of  those  four  annotations.   The  mean  inter-annotation correlation is 0.63 for valence and 0.60 for arousal. Let us note here that all subjects present in each video have been annotated. Figure \ref{va_annot} shows the 2D Valence-Arousal histogram of annotations of Aff-Wild2.

Aff-Wild2 is split into three subsets: training, validation and test. Partitioning is done in a subject independent manner, in the sense that a person can  appear  only  in  one  of  those  three  subsets. The  resulting  training, validation and test subsets consist of 346, 68 and 131 videos, respectively. 

\subsection{Aff-Wild2: Seven Basic Expression Annotation}

For the purposes of this Challenge, we build upon the former  Aff-Wild2's annotated part, for providing annotation in terms of the seven basic expressions; we annotate in total $539$ videos consisting of $2,595,572$ frames with $431$ subjects, $265$ of which are male and $166$ female. Eight  of  the  videos  display  two  subjects (all of which have been annotated). 


Seven experts performed the annotation of Aff-Wild2 for the seven basic expressions in a frame-by-frame basis; a platform-tool was developed in order to split each video into frames and let the experts annotate each videoframe. Let us mention that in this platform-tool, an expert could score a videoframe as having either one of the seven basic expressions or none (since there are affective states other than the seven basic expressions). Let us note again that all subjects appearing in each video have been annotated.

Due to subjectivity of annotators and wide ranging levels of images’ difficulty, there were some disagreements among annotators. We decided to keep only the annotations on which at least five (out of seven) experts agreed. Table \ref{expr_distr} shows the distribution of the seven basic expression annotations of Aff-Wild2.

\begin{table}[!h]
\caption{ Number of Annotated Images in Each of the Seven Basic Expressions   }
\label{expr_distr}
\centering
\begin{tabular}{ |c||c| }
\hline
Basic Expression & No of Images \\
\hline
\hline
Neutral & 1,268,631  \\
 \hline
Anger & 51,837  \\
 \hline
Disgust & 32,258  \\
 \hline
Fear &  27,388 \\
 \hline
Happiness & 389,517  \\
 \hline
Sadness & 172,612  \\
 \hline
Surprise & 99,391  \\
 \hline
\end{tabular}
\end{table}

Aff-Wild2 is split into three subsets: training, validation and test. Partitioning is done in a subject independent manner. The  resulting  training, validation and test subsets consist of 253, 71 and 223 videos, respectively. 

\subsection{Aff-Wild2: Eight Action Unit Annotation}

Aff-Wild2 has been partly annotated in terms of eight action units, as well. The annotated part consists of $56$ videos, with 63 subjects (32 males and 31 females), in $398,835$ frames. Seven of these videos display two subjects (both have been annotated). 

Three experts performed the annotation of Aff-Wild2 for the occurrence of eight action units in a frame-by-frame basis; a platform-tool (similar to the one used for annotating the seven basic expressions) was developed in order to split each video into frames and let the experts annotate each videoframe. The agreement between the annotators has not always been 100\%. Therefore, we decided to keep the annotations, on which all three experts agreed. Let us also note that all subjects present in each video have been annotated. Table \ref{au_distr} shows the name of the eight action units that have been annotated, the action that they are associated with and the distribution of their annotations in Aff-Wild2.

\begin{table}[h]
    \centering
        \caption{Distribution of AU annotations in Aff-Wild2}
    \label{au_distr}
\begin{tabular}{|c|c|c|}
\hline
  Action Unit \# & Action  &\begin{tabular}{@{}c@{}} Total Number \\  of Activated AUs \end{tabular} \\   \hline    
    \hline    
   AU 1 & inner brow raiser & 86,677 \\   \hline    
   AU 2 & outer brow raiser & 4,166 \\   \hline    
   AU 4 & brow lowerer & 56,327 \\  \hline    
   AU 6 & cheek raiser & 25,226 \\  \hline    
   AU 12 & lip corner puller & 35,675 \\  \hline    
   AU 15 & lip corner depressor & 3,340 \\  \hline    
   AU 20 & lip stretcher & 5,695 \\  \hline    
   AU 25 & lips part & 9,048 \\  \hline    
\end{tabular}
\end{table}

Aff-Wild2 is split into three subsets: training, validation and test. Partitioning is done in a subject independent manner. The  resulting  training, validation and test subsets consist of 37, 7 and 12 videos, respectively.

\subsection{Aff-Wild2 Pre-Processing: Cropped \& Cropped-Aligned Images} \label{pre-process}


At first, we split all videos into images (frames). Then, the SSH detector \cite{najibi2017ssh} based on the ResNet \cite{he2016deep} and trained on the WiderFace dataset \cite{yang2016wider} was used to extract face bounding boxes from all the images. The cropped images according to these bounding boxes were provided to the participating teams.
Also, 5 facial landmarks (two eyes, nose and two mouth corners) were extracted and used to perform similarity transformation (for face alignment \cite{avrithis2000broadcast}). The resulting cropped and aligned images were additionally provided to the participating teams. Finally, the cropped and aligned images were utilized in our baseline experiments, described in Section \ref{baseline}.

\section{Evaluation Metrics Per Challenge}\label{metrics}

Next, we present the metrics that will be used for assessing the performance of the developed methodologies of the participating teams in each Challenge.

\subsection{Valence-Arousal Estimation Challenge} 

\noindent The Concordance Correlation Coefficient (CCC) is widely used in measuring the performance of dimensional emotion recognition methods, such as in the series of AVEC challenges \cite{ringeval2019avec}. CCC evaluates the agreement between two time series (e.g., all video annotations and predictions) by scaling their correlation coefficient with their mean square difference. In this way, predictions that are well correlated with the annotations but shifted in value are penalized in proportion to the deviation. CCC takes values in the range $[-1,1]$, where $+1$ indicates perfect concordance and $-1$ denotes perfect discordance. The highest the value of the CCC the better the fit between annotations and predictions, and therefore high values are desired.
CCC is defined as follows:

\begin{equation} \label{ccc}
\rho_c = \frac{2 s_{xy}}{s_x^2 + s_y^2 + (\bar{x} - \bar{y})^2},
\end{equation}

\noindent
where $s_x$ and $s_y$ are the variances of all video valence/arousal annotations and predicted values, respectively, $\bar{x}$ and $\bar{y}$ are their corresponding mean values and $s_{xy}$ is the corresponding covariance value.

The mean value of CCC for valence and arousal estimation will be adopted as the main evaluation criterion. 

\begin{equation} \label{va}
\mathcal{E}_{total} = \frac{\rho_a + \rho_v}{2},
\end{equation}

\subsection{Seven Basic Expression Classification Challenge}\label{evaluation}

\noindent The $F_1$ score is a weighted average of the recall (i.e., the ability of the classifier to find all the positive samples) and precision (i.e., the ability of the classifier not to label as positive a sample that is negative). The $F_1$ score reaches its best value at 1 and its worst score at 0. The $F_1$ score is defined as:

\begin{equation} \label{f1}
F_1 = \frac{2 \times precision \times recall}{precision + recall}
\end{equation}

The $F_1$ score for emotions is computed based on a per-frame prediction (an emotion category is specified in each frame).

Total accuracy (denoted as $\mathcal{T}Acc$) is defined on all test samples and is the  fraction of predictions that the model got right. Total accuracy reaches its best value at 1 and its worst score at 0. It is defined as:

\begin{equation} \label{total accuracy}
\mathcal{T}Acc = \frac{\text{Number of Correct Predictions}}{\text{Total Number of Predictions}}
\end{equation}

A weighted average between the $F_1$ score and the total accuracy, $\mathcal{T}Acc$, will be the main evaluation criterion:

\begin{equation} \label{expr}
\mathcal{E}_{total} = 0.67 \times F_1 + 0.33 * \mathcal{T}Acc,
\end{equation}

\subsection{Eight Action Unit Detection Challenge}\label{evaluation2}

To obtain the overall score for the AU detection Challenge,
we first obtain the $F_1$ score for each AU independently, and then compute the (unweighted) average over all 8 AUs (denoted as $\mathcal{A}F_1$) :
\begin{equation} \label{au1}
\mathcal{A}F_1 = \sum_{i=1}^{8} F_1^i
\end{equation}

The $F_1$ score for AUs is computed based on a per-frame detection (whether each AU is present or absent). 

The average between the $\mathcal{A}F_1$ score and the total accuracy, $\mathcal{T}Acc$, will be the main evaluation criterion:

\begin{equation} \label{aus}
\mathcal{E}_{total} = 0.5 \times \mathcal{A}F_1 + 0.5 * \mathcal{T}Acc
\end{equation}

\section{Baseline \& Participating Teams' Systems and Results} \label{baseline}

All baseline systems rely exclusively on existing open-source machine learning toolkits to ensure the reproducibility of the results.
In this Section, we first describe the baseline systems developed for each Challenge, then we present the participating teams's algorithms that ranked in the top-3 of each Challenge and finally report their obtained results. 

At first, let us mention that we utlized the cropped and aligned images from Aff-Wild2, as described in Section \ref{pre-process}. These images are then resized to dimension $96 \times 96 \times 3$. The pixel intensities are normalized to take values in [-1,1]. No on-the-fly or off-the-fly data augmentation technique \cite{kuchnik2018efficient,kollias2018photorealistic,kolliasijcv} was utilized.


\subsection{Baseline System: Valence-Arousal Estimation Challenge}

The architecture that was used for estimating valence and arousal was based on that of PatchGAN \cite{isola2017image,zhu2017unpaired,choi2018stargan}. PatchGAN is a deep convolutional neural network initially designed to classify patches of an input image, rather than the entire image, as real or fake.
The PatchGAN was the discriminator of the pix2pix architecture \cite{isola2017image}. The output of the network is a single feature map of real/fake predictions that was averaged to give a single score. In StarGAN \cite{choi2018stargan}, PatchGAN was additionally used as a classifier. Here, we adopt PatchGAN for valence-arousal regression. The exact architecture used, can be seen in Table \ref{patchgan}. It was implemented in TensorFlow, trained from scratch, for around two days on a Titan X GPU, with a learning rate of $10^{-4}$. PatchGAN's implementation code is publicly available from groups that implemented pix2pix, CycleGAN \cite{zhu2017unpaired} and StarGAN.

\begin{table}[!h]
\centering
\caption{PatchGAN adopted for valence-arousal estimation. Leaky Rely follows each convolutional layer.}
\label{patchgan}
\scalebox{1.}{
\begin{tabular}{|c|c|c|}
\hline
Name & Type & Filter \\
\hline
\hline
conv  & weights & (4, 4, 3, 64) \\ \hline
conv 1 & weights & (4, 4, 64, 128) \\ \hline
conv 2 & weights & (4, 4, 128, 256) \\ \hline
conv 3 & weights & (4, 4, 256, 512) \\ \hline
conv 4 & weights & (4, 4, 512, 1024) \\ \hline
conv 5 & weights & (4, 4, 512, 1024) \\ \hline
conv 6 & weights & (4, 4, 1024, 2048) \\ \hline
conv 7 & weights/D-label & (1, 1, 2048, 2) \\ 
\hline
\end{tabular}}
\end{table}



\subsection{Baseline System: Seven Basic Expression Classification \& Eight Action Unit Detection Challenges}

The architectures that were used for the tasks of classification into the seven basic expressions and detection of eight action units, were based on the architecture of MobileNetV2 \cite{sandler2018mobilenetv2}. MobileNetV2 belongs to the class of efficient models called MobileNets \cite{howard2017mobilenets} that are light-weight deep neural networks. They are based on a streamlined architecture that uses depth-wise separable convolutions which dramatically reduce the complexity, cost and model size of the network. For more details regarding this class of architectures and the MobileNetV2 network, we refer the interested reader to \cite{sandler2018mobilenetv2}. Table \ref{mobilenetv2} shows the basic structure of MobileNetV2.

\begin{table}[!h]
\centering
\caption{The MobileNetV2 network}
\label{mobilenetv2}
\scalebox{1.}{
\begin{tabular}{|c|c|c|}
\hline
Name & Type & Filter \\
\hline
\hline
conv &  weights &  (3, 3, 3, 32) \\ \hline
expanded conv &  depthwise &  (3, 3, 32, 1) \\ \hline
expanded conv &  project &  (1, 1, 32, 16) \\ \hline
expanded conv 1 &  expand &  (1, 1, 16, 96) \\ \hline
expanded conv 1 &  depthwise &  (3, 3, 96, 1) \\ \hline
expanded conv 1 &  project &  (1, 1, 96, 24) \\ \hline
expanded conv 2 &  expand &  (1, 1, 24, 144) \\ \hline
expanded conv 2 &  depthwise &  (3, 3, 144, 1) \\ \hline
expanded conv 2 &  project &  (1, 1, 144, 24) \\ \hline
expanded conv 3 &  expand &  (1, 1, 24, 144) \\ \hline
expanded conv 3 &  depthwise &  (3, 3, 144, 1) \\ \hline
expanded conv 3 &  project &  (1, 1, 144, 32) \\ \hline
expanded conv 4 &  expand &  (1, 1, 32, 192) \\ \hline
expanded conv 4 &  depthwise &  (3, 3, 192, 1) \\ \hline
expanded conv 4 &  project &  (1, 1, 192, 32) \\ \hline
expanded conv 5 &  expand &  (1, 1, 32, 192) \\ \hline
expanded conv 5 &  depthwise &  (3, 3, 192, 1) \\ \hline
expanded conv 5 &  project &  (1, 1, 192, 32) \\ \hline
expanded conv 6 &  expand &  (1, 1, 32, 192) \\ \hline
expanded conv 6 &  depthwise &  (3, 3, 192, 1) \\ \hline
expanded conv 6 &  project &  (1, 1, 192, 64) \\ \hline
expanded conv 7 &  expand &  (1, 1, 64, 384) \\ \hline
expanded conv 7 &  depthwise &  (3, 3, 384, 1) \\ \hline
expanded conv 7 &  project &  (1, 1, 384, 64) \\ \hline
expanded conv 8 &  expand &  (1, 1, 64, 384) \\ \hline
expanded conv 8 &  depthwise &  (3, 3, 384, 1) \\ \hline
expanded conv 8 &  project &  (1, 1, 384, 64) \\ \hline
expanded conv 9 &  expand &  (1, 1, 64, 384) \\ \hline
expanded conv 9 &  depthwise &  (3, 3, 384, 1) \\ \hline
expanded conv 9 &  project &  (1, 1, 384, 64) \\ \hline
expanded conv 10 &  expand &  (1, 1, 64, 384) \\ \hline
expanded conv 10 &  depthwise &  (3, 3, 384, 1) \\ \hline
expanded conv 10 &  project &  (1, 1, 384, 96) \\ \hline
expanded conv 11 &  expand &  (1, 1, 96, 576) \\ \hline
expanded conv 11 &  depthwise &  (3, 3, 576, 1) \\ \hline
expanded conv 11 &  project &  (1, 1, 576, 96) \\ \hline
expanded conv 12 &  expand &  (1, 1, 96, 576) \\ \hline
expanded conv 12 &  depthwise &  (3, 3, 576, 1) \\ \hline
expanded conv 12 &  project &  (1, 1, 576, 96) \\ \hline
expanded conv 13 &  expand &  (1, 1, 96, 576) \\ \hline
expanded conv 13 &  depthwise &  (3, 3, 576, 1) \\ \hline
expanded conv 13 &  project &  (1, 1, 576, 160) \\ \hline
expanded conv 14 &  expand &  (1, 1, 160, 960) \\ \hline
expanded conv 14 &  depthwise &  (3, 3, 960, 1) \\ \hline
expanded conv 14 &  project &  (1, 1, 960, 160) \\ \hline
expanded conv 15 &  expand &  (1, 1, 160, 960) \\ \hline
expanded conv 15 &  depthwise &  (3, 3, 960, 1) \\ \hline
expanded conv 15 &  project &  (1, 1, 960, 160) \\ \hline
expanded conv 16 &  expand &  (1, 1, 160, 960) \\ \hline
expanded conv 16 &  depthwise &  (3, 3, 960, 1) \\ \hline
expanded conv 16 &  project &  (1, 1, 960, 320) \\ \hline
conv 1 & weights &  (1, 1, 320, 1280) \\ 
\hline
\end{tabular}}
\end{table}

Let us note that: i) batch normalization is applied after each convolutional or expanded convolutional layer, ii) the non-linearity is Relu6 and iii) no average pooling is conducted in the end. After the final convolutional layer (shown in Table \ref{mobilenetv2}), a fully connected layer follows (with 7 units if the task is to predict the 7 basic expressions, or 8 units if the task is to detect the 8 action units) and on top of that is a softmax or sigmoid layer, respectively.

MobileNetV2 was implemented in TensorFlow, trained from scratch, for around three days on a Titan X GPU, with a learning rate of $10^{-4}$. MobileNetV2 is released as part of tf-slim library.

\subsection{Top-3 Performing Teams per Challenge and their Methodologies}

At first let us mention that in total sixty two (62) different teams registered for this Competition. Unfortunately due to the coronavirus a lot of these groups notified that they would not be able to send the test results before the final deadline for the Competition.
Next, we present five methodologies that displayed the best performances in each Challenge and ranked in the top-3. 

The NISL2020 team \cite{deng2020fau} participated in all three Challenges and ranked in the first, third and first place of the Valence-Arousal Estimation, Seven Basic Expression Classification and Eight Action Unit Detection Challenges, respectively. 
Their methodology tackled multi-task learning of these three tasks. The authors proposed an algorithm for their developed multi-task model to learn from partial labels. At first, they trained a teacher model to perform all three tasks, where each instance is trained by the ground truth label of its corresponding task. Then, they used the outputs of the teacher model and the ground truths to train the student model, resulting in the latter outperforming the teacher model. Finally, ensemble methodologies were used to further boost the performance of the model.

The TNT team \cite{kuhnke2020two} participated in all three Challenges and ranked in the second, first and second place of the Valence-Arousal Estimation, Seven Basic Expression Classification and Eight Action Unit Detection Challenges, respectively.
Their methodology tackled multi-task learning of these three tasks using multi-modal (images and audio) input. The authors proposed a two-stream aural-visual analysis model in which audio and image streams are first processed separately and fed into a convolutional neural
network. The authors did not use  recurrent architectures for temporal analysis but used instead  temporal convolutions. Furthermore, the model was given access to additional features extracted during face-alignment pre-processing. At training time, correlations between different emotion representations are exploited so as to improve the model's performance.

The ICT-VIPL-VA team \cite{zhang2020m}  participated in the Valence-Arousal Estimation Challenge and ranked in the third place. 
Their methodology fused both visual features extracted from videos and acoustic features extracted from audio tracks. 
To extract the visual features, the authors followed a CNN-RNN paradigm, in which spatio-temporal visual features \cite{rapantzikos2007bottom} are extracted with a 3D convolutional network and / or a pretrained 2D convolutional network, and are fused through a bidirectional recurrent neural network. The audio features are extracted from a GRU-MLP network.

The ICT-VIPL-Expression team \cite{liu2020emotion}  participated in the Seven Basic Expression Classification Challenge and ranked in the second place.
Their methodology combined a Deep Residual Network with convolutional block attention module and also Bidirectional Long-Short-Term Memory Units. The authors also visualized the learned attention maps and analyzed the importance of different regions in facial expression recognition.

The SALT team \cite{pahl2020multi} participated in the Eight Action Unit Detection Challenge and ranked in the third place.
Their methodology included a multi-label class balancing algorithm as a pre-processing step for overcoming the imbalanced occurrences of Action Units in
the training dataset. Then a ResNet was trained using the augmented training dataset.

\subsection{Results}

Table \ref{ccc_results} presents the CCC evaluation of valence and arousal predictions on the Aff-Wild2 test (validation) set, of the baseline network (PatchGAN) and the networks developed by the top-3 performing teams of this Challenge. It can be observed that the NISL2020 team achieved the best overall score and the best arousal CCC, whereas the TNT team achieved the best valence CCC. It should be noted that all participating teams outperformed the baseline by a large margin.

\begin{table}[!h]
\caption{Baseline results for VA estimation on the test (validation) set of Aff-Wild2 of: i) the baseline network (PatchGAN) and ii) the networks developed by the top-3 performing teams; '-' means that no result is reported in the corresponding paper; $\mathcal{E}_{total}$ is the mean valence and arousal CCC}
\label{ccc_results}
\centering
\begin{tabular}{ |c||c|c|c| }
\hline
\multicolumn{1}{|c||}{Team}  & \multicolumn{2}{c|}{CCC} & \multicolumn{1}{c|}{$\mathcal{E}_{total}$} \\
\hline
      & Valence & Arousal &   \\
 \hline
 \hline
NISL2020 & 0.44 (0.335) & \textbf{0.454} (0.515) & \textbf{0.447} (0.425) \\
\hline
TNT & \textbf{0.448} (-) & 0.417 (-) & 0.433 (-) \\
\hline
ICT-VIPL-VA & 0.361 (0.32) & 0.408 (0.55) & 0.385 (0.435) \\
\hline
\hline
Baseline  & 0.11 (0.14) & 0.27 (0.24) & 0.19 (0.19)  \\
 \hline
\end{tabular}
\end{table}

Table \ref{expr_results} presents the performance on the test (validation) set of Aff-Wild2, of the baseline network (MobileNetV2), and of the networks developed by the top-3 performing teams of the  Seven Basic Expression Classification Challenge. The performance metric is a weighted average between the F1 score and the total accuracy, as discussed in Section \ref{evaluation}. It can be observed that the TNT team outperformed by a large margin all other teams in both F1 Score and Total accuracy (and thus in the total evaluation metric). Let us mention that all participating teams outperformed the baseline.

\begin{table}[!h]
\caption{Baseline results for basic expression classification on the test (validation) set of Aff-Wild2 of: i) the baseline network (MobileNetV2) and ii) the networks developed by the top-3 performing teams; '-' means that no result is reported in the corresponding paper; $\mathcal{E}_{total} = 0.67 \times F_1 + 0.33 * \mathcal{T}Acc$}
\label{expr_results}
\scalebox{.95}{
\centering
\begin{tabular}{ |c||c|c|c| }
\hline
\multicolumn{1}{|c||}{Team}  & \multicolumn{1}{c|}{\begin{tabular}{@{}c@{}}F1 \\ Score \end{tabular}} & \multicolumn{1}{c|}{\begin{tabular}{@{}c@{}}Total \\  Accuracy \end{tabular}} & \multicolumn{1}{c|}{$\mathcal{E}_{total}$} \\
 \hline
 \hline
TNT & \textbf{0.398} (-) &  \textbf{0.734} (-) &  \textbf{0.509} (-) \\
\hline
ICT-VIPL-Expression & 0.286 (0.333) & 0.655 (0.64) & 0.408 (0.434) \\
\hline
NISL2020 & 0.27 (-) & 0.68 (-) & 0.405 (0.493) \\
\hline
\hline
Baseline & 0.15 (0.21) & 0.605 (0.664) & 0.30 (0.36)  \\  
 \hline
\end{tabular}
}
\end{table}

Table \ref{aus_results} presents the performance on the test (validation) set of Aff-Wild2, of the baseline network (MobileNetV2), and of the networks developed by the top-3 performing teams of the  Eight Action Unit Detection Challenge. The performance metric is the average between the F1 score and the total accuracy, as discussed in Section \ref{evaluation2}. It can be seen that the NISL2020 team achieved the best performance in the average F1 Score, whereas the TNT team achieved the best one in the Total Accuracy. The NISL2020 team achieved a better overall performance, with a very small difference, than the TNT one. Again all participating teams outperformed the baseline.

\begin{table}[!h]
\caption{Baseline results for action unit detection on the test (validation) set of Aff-Wild2 of: i) the baseline network (MobileNetV2) and ii) the networks developed by the top-3 performing teams; '-' means that no result is reported in the corresponding paper; $\mathcal{E}_{total} = 0.5 \times \mathcal{A}F_1 + 0.5 * \mathcal{T}Acc$}
\label{aus_results}
\centering
\begin{tabular}{ |c||c|c|c| }
\hline
\multicolumn{1}{|c||}{Team}  & \multicolumn{1}{c|}{\begin{tabular}{@{}c@{}}Average \\ F1 Score \end{tabular}} & \multicolumn{1}{c|}{\begin{tabular}{@{}c@{}}Total \\  Accuracy \end{tabular}} & \multicolumn{1}{c|}{$\mathcal{E}_{total}$} \\
 \hline
 \hline
NISL2020 & \textbf{0.309} (-) & 0.905 (-) & \textbf{0.607} (0.591) \\
 \hline
TNT & 0.27 (-) &  \textbf{0.932} (-) &  0.601 (-) \\
\hline
SALT & 0.216 (0.24) & 0.886 (-) & 0.551 (0.59) \\
\hline
\hline
Baseline & 0.16 (0.22)  & 0.36 (0.4)  & 0.26 (0.31)  \\
 \hline
\end{tabular}
\end{table}

\section{Conclusion}\label{conclusion}
In this paper we have presented the  Affective Behavior Analysis in-the-wild Competition (ABAW) 2020.  It comprises  three Challenges targeting: i) valence-arousal estimation, ii) seven basic expression classification and iii) eight action unit detection. The database utilized for this
Competition has been derived from the Aff-Wild2, the large-scale and first  database annotated for all these three behavior tasks.
We have also presented the baseline networks and their results. Finally, we described the top-3 performing, per Challenge, team methodologies and presented their results. 

\bibliographystyle{ieee}
\bibliography{egbib}

\end{document}